# GENERATING THE CLOUD MOTION WINDS FIELD FROM SATELLITE CLOUD IMAGERY USING DEEP LEARNING APPROACH

Chao Tan

College of Computer Science and Engineering, Chongqing University of Technology
istvartan@outlook.com

**ABSTRACT**

*Cloud motion winds (CMW) are routinely derived by tracking features in sequential geostationary satellite infrared cloud imagery. In this paper, we explore the cloud motion winds algorithm based on data-driven deep learning approach, and different from conventional hand-craft feature tracking and correlation matching algorithms, we use deep learning model to automatically learn the motion feature representations and directly output the field of cloud motion winds. In addition, we propose a novel large-scale cloud motion winds dataset (CMWD) for training deep learning models. We also try to use a single cloud imagery to predict the cloud motion winds field in a fixed region, which is impossible to achieve using traditional algorithms. The experimental results demonstrate that our algorithm can predict the cloud motion winds field efficiently, and even with a single cloud imagery as input.*

*Index Terms*—Cloud motion winds, Dense optic flow prediction, Deep convolutional neural network

## 1. INTRODUCTION

Cloud motion winds (CMW) are derived from geostationary satellite-observed motions of clouds features, which have been shown to provide reasonable estimates of the ambient tropospheric wind [17]. The applications of cloud motion winds are wide ranging. They provide crucial information over wide regions of the Southern Hemisphere which void of conventional ground observations and contribute to forecast diagnostics of weather conditions, as well as numerical weather prediction (NWP). In addition, cloud motion winds field can also provide basic guidance for such field as solar energy prediction [14]. The traditional cloud motion winds algorithm first generates a motion vector for each pixel using feature tracking and correlation matching methods, and then revises the vector field through a series of post-processing operations [11-13,18]. Among them, Lu [15] propose a new tracer selection procedure named temporal difference method to calculate the low-level atmosphere motion vectors. Bresky [16] develop a new approach that isolates the dominant local motion within a cloud scene and minimizes the smoothing of the motion estimate. Cloud motion winds algorithms have advanced in accuracy from the mid-1960s, but they usually require much computing time, especially when it comes to capturing dense cloud motion winds field at high resolution.

The CMW algorithm is similar but not identical to the task of dense optical flow estimation in computer vision. In this paper, we will learn from this idea and use a deep learning model to predict the vector field of cloud motion winds. The reason for using deep learning approach is that it can automatically learn the representation of motion features from a large amount of data samples, and it can also greatly improve the calculation speed and efficiency of the algorithm. Many existing deep unsupervised learning based optical flow estimation methods constrain the motion between two frames by minimizing photometric and smoothing loss [3-6]. Among them, Ren [1] devise a simple yet effective unsupervised method for learning optical flow, by directly minimizing photometric consistency. Meister [2] inspired by classical energy-based optical flow methods and design an occlusion-aware bidirectional dense flow estimation and robust census transform based unsupervised loss to evade the need for ground truth flow.

In this paper, we use a fully convolutional deep network to generate the CMW field between two geostationary satellite cloud imagery sequences. However, duo to the uncertainty of atmospheric movement, the brightness consistency between two adjacent satellite cloud imageries is not guaranteed, so the unsupervised optic flow estimation based on photometric consistency is not suitable for use. Therefore, in order to use deep neural network to predict the CMW field, we use a more robust PCAFlow [8] to generate rough labels for the training samples in advance, and then use them as the supervision of the deep neural network for training. At the same time, in order to train our deep learning model, we propose a novel large-scale cloud motion winds dataset named CMWD, in which each sample contains two geostationary satellite cloud imagery at fixed intervals.

We also try to use only one frame of satellite cloud imagery to estimate the cloud motion winds filed. This requires the model to learn the characteristics of atmospheric movement from dataset and to have sufficient generalization ability. It should be noted that this task is not feasible for other existing CMW algorithms. However, since data-driven deep neural network can learn the hidden patterns of atmosphere motion from the dataset (for example, typhoons in the Pacific North-west can only rotate counterclockwise but not clock wise), we can use one satellite cloud imagery to make appropriate estimates of cloud motion winds field. To our best knowledge, we are the first to use deep learning to conduct research on

project page: https://chao-tan.gitee.io/projects/cmw-net/project-page.html

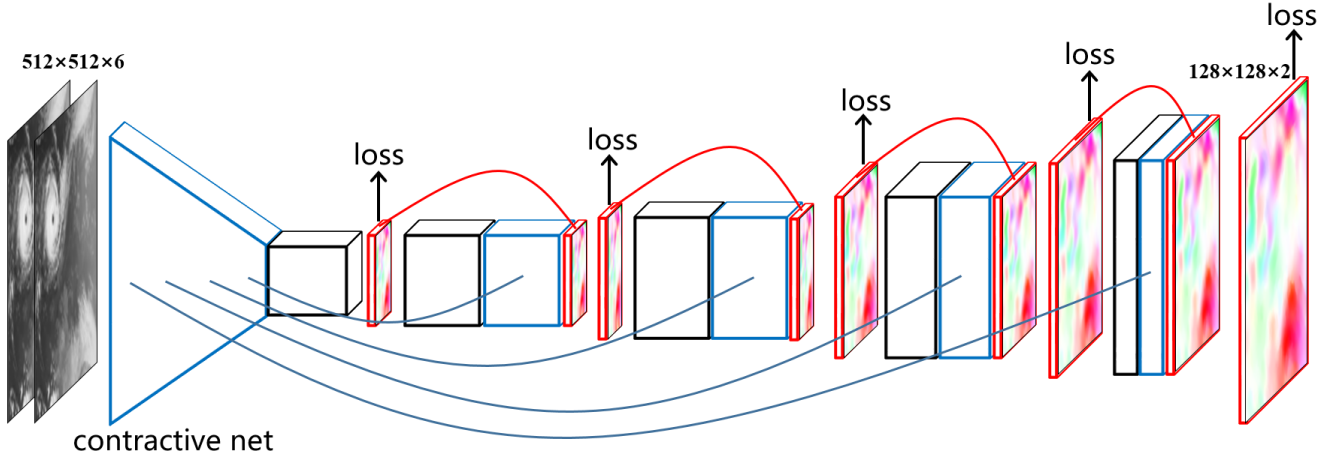

**Figure 1**. The network architecture of FlowNetS model. Two consecutive geostationary meteorological satellite cloud imageries are taken as input, and a cloud motion winds vector field is generated using a multi-stage refinement process. Feature maps from the contractive part, as well as intermediate motion vector field predictions, are used in the "up convolutional" part.

CMW filed generation, which also provides resources for the application of deep learning in the field of weather prediction.

## 2. DATASET AND METHODOLOGY

### 2.1. Cloud Motion Winds Dataset (CMWD)

In order to train our data-driven based deep learning model, we propose a novel large-scale cloud motion winds dataset named CMWD, which contains 6388 training samples and 715 samples for testing from Helianthus 8 high-precision geostationary infrared meteorological satellite. Each sample consists of two geostationary satellite cloud imageries with the size of 512 ×512, taken a one-hour intervals over a vast of the Pacific Northwest. In addition, we use the traditional unsupervised optical flow method to generate coarse labels for the training data, which can be easily implemented using many existing frameworks [8,10]. It should be noted that, like many existing works [7,9], in order to visualize the output optical flow, we encode the generated two-dimensional flow field with color in the HSV color model. Specifically, flow direction is encoded with color and magnitude with color intensity and white corresponds to no motion. To our best knowledge, CMWD is the first large-scale cloud motion winds dataset for deep learning research, which can provide benchmark for subsequent deep learning research in the field of meteorological research and prediction.

### 2.2. Proposed Network

In this paper, we will use a data-driven deep learning model to estimate the cloud motion winds field. Specifically, we want to learn a deep convolutional network, which maps two adjacent geostationary infrared satellite cloud imageries into two-dimensional cloud motion winds field. It should be noted that, as describe in Section 2.1, we use color coding to represent the two dimensional flow, which allows us to visualize the flow field intuitively. In other words, given an RGB image pair as input, $x \in R^{6 \times h \times w}$, our model needs to learn a non-linear transformation to the corresponding cloud motion winds flow field, $y \in R^{2 \times h \times w}$, where $h$ and $w$ denote the height and width of input satellite cloud imagery respectively.

We use FlowNetS [7] as a reference network and the network structure is illustrated in Figure 1. This architecture consists of a contractive part followed by an expanding part. The contractive part takes as input two infrared cloud imageries stacked together, and processes them with a cascade of stride deconvolution layers. The expanding part implements a "skip layer" architecture that combines information from various levels of contractive part with "up convolving" layers to iteratively refine the coarse cloud motion winds filed predictions. In addition, it can be seen from Figure 1 that we use a loss compose of a final loss and several intermediate losses placed at various stages of the expansionary part. The intermediate losses are meant to guide earlier layers more directly towards the final objective. Specifically, we use endpoint error (EPE) as the supervised training loss, and its ground truth motion field is generated by PCAFlow [8]. It should be noted that in order to make the generated cloud motion winds field as smooth as possible, the image size of the final output motion vector field of our model is a quarter of the input image size, and then the flow field with the same width and height as the input image is obtained via the nearest neighbor interpolation.

### 2.3. Implement details

We use the Adam solver [19] with a basic learning rate of 0.001 and the first and second momentum values are 0.5 and 0.999 respectively to optimize our network. The size of input infrared cloud imagery is 512×512 and the output cloud motion winds field of our model is 128×128. It should be noted that we do not use data argumentation strategies such as random flipping, because we believe that they will affect the influence the model to learn the motion patterns of the atmosphere. We adopt an endpoint error (EPE) loss compose of a final loss and four intermediate losses placed at different stages of the expansionary part. Like [7], we set the weights of the five losses from bottom to top layers to 0.005, 0.01, 0.02, 0.08 and 0.32 respectively. We train our network in a total of 80 epochs with a mini-batch size of 8, and we reduce the learning rate from 0.001 to 0.0001 after 50 epochs.

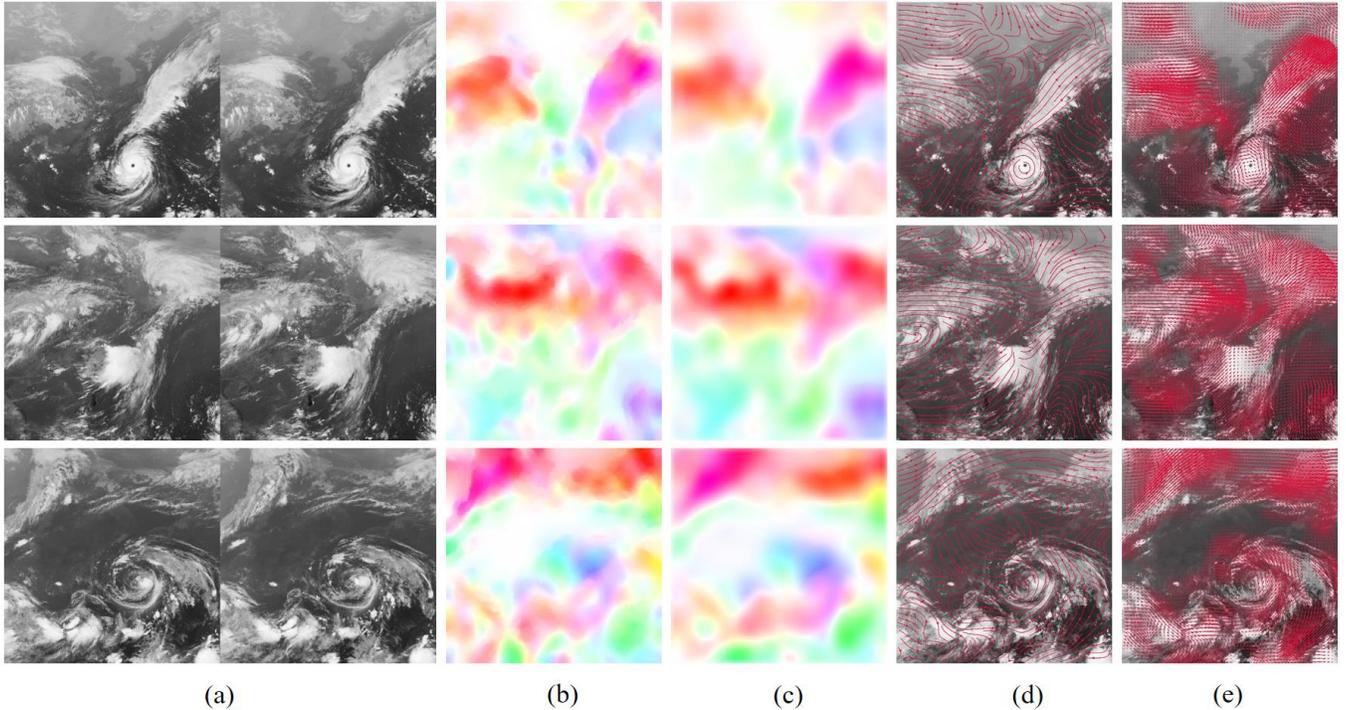

**Figure 2**. Three experimental results of test samples on CMWD. (a) is the input two infrared cloud imageries. (b) and (c) are the ground truth and predicted visual flow field maps respectively. (d) is the generated flux streamline diagram, and (e) is the dense vector arrow diagram.

## 3. EXPERIMENTS

### 3.1. Experiments on Paired Cloud Imagery

We first use the paired infrared satellite cloud imageries on CMWD dataset to conduct the experiment of cloud motion winds field estimation. As mentioned earlier in this paper, we visualize the two-dimensional cloud motion winds field as an RGB image, so that we can intuitively compare the fitting effect of the model. At the same time, we also convert the generated vector field into two visualization products, one is the common dense vector arrow diagram. The other is the flux streamline diagram, which can continuously display the direction and intensity of the wind field. The experimental results of some testing samples are shown in Figure 2. It should be noted that for the vector arrow diagram (the last row in Figure 2), we enlarge the wind arrow five times and automatically filter out the small arrows for a better visual effect. It can be seen from Figure 2 that our model can well estimate the upper-air wind motion at its location, weather it is a large air mass movement or a small-to-medium-scale anticyclonic spiral movement. As can be seen from the second and third columns of Figure 2, due to the complex fitting capabilities of our proposed model, we can use a more efficient deep neural network to fit the features movement of the input satellite cloud imagery pair. And since the rough label of the sample contains noise, we finally use the nearest neighbor interpolation to obtain a smoother result. In addition, since the neural network does not perform feature matching and other operations on a pixel-by-pixel basis, it can process high-resolution input images more efficiently. Specifically, with the input si-ze of 512×512, our model is more than 12 times faster than the traditional PCAFlow [8] in the inference stage.

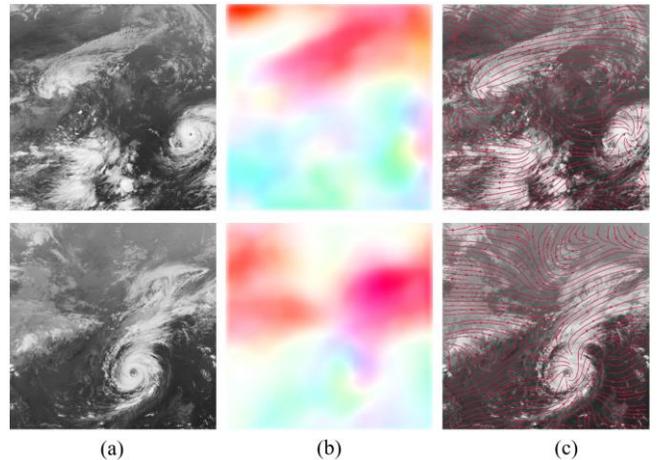

**Figure 3**. Two experimental results on CMWD. (a) is the input cloud imagery. (b) is the predicted visual flow field map and (c) is the generated flux streamline diagram.

### 3.2. Experiments on Single Cloud Imagery

We also try to use single infrared cloud iamgery to estimate the CMW field, which is not feasible for conventional CMW algorithms. However, because deep neural network can learn from a large amount of data, many basic patterns of atmospheric motion (such as the high-latitude typhoon in the Northwest Pacific Ocean moving counterclockwise to the west or north in large cases) can be detected by the model, and then predict through the state of the whole cloud system. It should be noted that in this experiment, the input image is only an infrared cloud image, $x \in R^{3 \times h \times w}$, other settings are same as the experiment in Section 3.1. The experimental results of some test samples are shown in Figure 3. From the experimental results, it can be seen that even with only one input image, our model is still able to make a rough estimate of CMW field under the current situation. Meanwhile, our model can separate different cloud structures in cloud iamgery and make different estimates accordingly (such as continental air mass and typhoon structure). This experiment embodies the learning ability and performance of the data-driven deep learning method, and further illustrates the powerful ability of weather analysis and prediction based on artificial intelligence tools.

### 4. CONCIUSIONS

In this work, we propose to use a deep convolutional neural network to estimate the cloud motion winds field. Meanwhile, we also present a large-scale cloud motion winds dataset for deep learning research. The experimental results illustrate the efficient performance of our model, and it can also handle the case without paired satellite cloud iamgery input. In the future, we plan to use deep learning based model to process more analysis task of meteorological remote sensing data.

### REFERENCES

[1]. Ren Z, Yan J, Ni B, et al. Unsupervised deep learning for optical flow estimation[C]//Thirty-First AAAI Conference on Artificial Intelligence. 2017.
[2]. Meister S, Hur J, Roth S. Unflow: Unsupervised learning of optical flow with a bidirectional census loss[J]. arXiv preprint arXiv:1711.07837, 2017.
[3]. Zhu Y, Lan Z, Newsam S, et al. Hidden two-stream convolutional networks for action recognition [C]// Asian Conference on Computer Vision. Springer, Cham, 2018: 363-378.
[4]. Wang Y, Yang Y, Yang Z, et al. Occlusion aware unsupervised learning of optical flow[C]//Proceedings of the IEEE Conference on Computer Vision and Pattern Recognition. 2018: 4884-4893.
[5]. Liu P, King I, Lyu M R, et al. Ddflow: Learning optical flow with unlabeled data distillation[C]//Proceedings of the AAAI Conference on Artificial Intelligence. 2019, 33: 8770-8777.
[6]. Jason J Y, Harley A W, Derpanis K G. Back to basics: Unsupervised learning of optical flow via brightness constancy and motion smoothness[C]//European Conference on Computer Vision. Springer, Cham, 2016: 3-10.
[7]. Dosovitskiy A, Fischer P, Ilg E, et al. Flownet: Learning optical flow with convolutional networks [C]// Proceedings of the IEEE international conference on computer vision. 2015: 2758-2766.
[8]. Wulff J, Black M J. Efficient sparse-to-dense optical flow estimation using a learned basis and layers [C]//Proceedings of the IEEE Conference on Computer Vision and Pattern Recognition. 2015: 120-130.
[9]. Ilg E, Mayer N, Saikia T, et al. Flownet 2.0: Evolution of optical flow estimation with deep networks [C]//Proceedings of the IEEE conference on computer vision and pattern recognition. 2017: 2462-2470.
[10]. Weinzaepfel P, Revaud J, Harchaoui Z, et al. DeepFlow: Large displacement optical flow with deep matching [C]//Proceedings of the IEEE international conference on computer vision. 2013: 1385-1392.
[11]. Schmetz J, Holmlund K, Hoffman J, et al. Operational cloud-motion winds from Meteosat infrared images[J]. Journal of applied meteorology, 1993, 32(7): 1206-1225.
[12]. Wang Z, Sui X, Zhang Q, et al. Derivation of cloud-free-region atmospheric motion vectors from FY-2E thermal infrared imagery[J]. Advances in Atmospheric Sciences, 2017, 34(2): 272-282.
[13]. Kaur I, Deb S K, Kishtawal C M, et al. Low level cloud motion vectors from Kalpana-1 visible images[J]. Journal of Earth System Science, 2013, 122(4): 935-946.
[14]. Cros S, Liandrat O, Sebastien N, et al. Extracting cloud motion vectors from satellite images for solar power forecasting[C]//2014 IEEE Geoscience and Remote Sensing Symposium. IEEE, 2014: 4123-4126.
[15]. Yang L, Wang Z, Chu Y, et al. Water vapor motion window images for cloud-free regions: The temporal difference technique[J]. Advances in Atmospheric Sciences, 2014, 31(6): 1386-1394.
[16]. Bresky W C, Daniels J M, Bailey A A, et al. New methods toward minimizing the slow speed bias associated with atmospheric motion vectors[J]. Journal of applied meteorology and climatology, 2012, 51(12): 2137-2151.
[17]. Sears J, Velden C S. Validation of satellite-derived atmospheric motion vectors and analyses around tropical disturbances[J]. Journal of applied meteorology and climatology, 2012, 51(10): 1823-1834.
[18]. Lazzara M A, Dworak R, Santek D A, et al. High-latitude atmospheric motion vectors from composite satellite data[J]. Journal of Applied Meteorology and Climatology, 2014, 53(2): 534-547.
[19]. Kingma D P, Ba J. Adam: A method for stochastic optimization[J]. arXiv preprint arXiv:1412.6980, 201